\pdfoutput=1

\documentclass[11pt]{article}

\usepackage[]{acl}

\usepackage{times}
\usepackage{latexsym}

\usepackage[T1]{fontenc}

\usepackage[utf8]{inputenc}

\usepackage{microtype}

\usepackage{inconsolata}

\usepackage{times}
\usepackage{latexsym}

\usepackage{microtype}

\usepackage{graphicx}
\usepackage{subcaption}
\usepackage{amsmath}
\usepackage{amssymb}
\usepackage{booktabs}
\usepackage{multirow}
\usepackage{enumitem}
\usepackage{cleveref}

\usepackage{changepage}

\usepackage{tabularx}

\title{A Simple Baseline for Beam Search Reranking}

  \author{
Lior Vassertail \qquad Omer Levy\\
The Blavatnik School of Computer Science\\
Tel Aviv University
}

\begin{document}
\maketitle

\begin{abstract}
Reranking methods in machine translation aim to close the gap between common evaluation metrics (e.g. BLEU) and maximum likelihood learning and decoding algorithms.
Prior works address this challenge by training models to rerank beam search candidates according to their predicted BLEU scores, building upon large models pretrained on massive monolingual corpora -- a privilege that was never made available to the baseline translation model.
In this work, we examine a simple approach for training rerankers to predict translation candidates' BLEU scores without introducing additional data or parameters.
Our approach can be used as a clean baseline, decoupled from external factors, for future research in this area.
\end{abstract}

\section{Introduction}

Neural machine translation typically uses beam search decoding to generate multiple candidate translations from an underlying conditional language model.
These candidates are then \textit{reranked} using an alternative metric, usually their token-averaged log-probability, to select the output.
Previous works \cite{fernandes-etal-2022-quality, bhattacharyya-etal-2021-energy, lee-etal-2021-discriminative} show that training a reranking model to predict evaluation metrics scores such as BLEU \cite{papineni-etal-2002-bleu} can improve performance with respect to that metric.
However, these approaches provide the reranker with monolingual pretraining, back-translated data, or more parameters -- none of which are given to the baseline model.
This work proposes a simple method to train machine translation rerankers without introducing new data or enlarging the model.

We take the simple approach of fine-tuning the baseline translation model to predict the BLEU score of its own translations.
We sample training-set translations from the model's distribution using greedy search, beam search, top-$k$ sampling \cite{fan-etal-2018-hierarchical}, nucleus (top-$p$) sampling \cite{Holtzman2020TheCC}, and the ground truth to create training examples for the reranker.
We study the effect of training the reranker on increasing sample sizes, across four datasets of varying languages and sizes: IWSLT'14 English to German, Arabic, and Russian \cite{Cettolo2015ReportOT}; WMT'16 Romanian to English \cite{bojar-etal-2016-findings}.

Experiments show that rerankers trained to predict BLEU, in our setting, perform only slightly better than the baseline token-averaged log-probability reranker.
Achieving this level of performance requires a reranking training set with dozens of generated translations for every source sentence.
When interpolating with the baseline reranking method, only a few generated translations are needed for the reranking training set to exceed the baseline's performance, but further enlarging the dataset yields only limited gains.
Further correlation analysis shows that even though the reranker's top choice results in minor improvements, it is significantly better than the baseline at ranking the entire list.

To the best of our knowledge, our work is the first to explore BLEU-driven beam search reranking in a controlled setting, without adding training data, model parameters, or other factors beyond the objective of predicting BLEU scores.
A natural question following to our findings is whether the information available from BLEU scores may be sufficient on its own to train rerankers that are substantially better than the standard token-averaged log-probability heuristic.
To test this hypothesis and identify the main factors leading to reranking improvements, we recommend using our BLEU-only approach as a baseline in future studies.

\section{Background: Reranking}

The current practice in machine translation is to use beam search to generate $k$ high-probability candidates from a conditional language model.
During search, each candidate prefix is scored according to its cumulative log-probability, i.e. the sum of log-probabilities that the model assigned to each decoded token:
$
\sum_i^{|y|} \log P_t(y_i|x, y_{<i})
$,
where $P_t$ denotes the \textit{translation} model.

A priori, higher scores should correlate with better candidates.
In practice, this metric is biased towards shorter sequences, since each additional token adds a negative number to the score.
The standard approach therefore \textit{reranks} the final $k$ candidates produced by beam search according to their \textit{token-averaged} log probability:
$
\frac{1}{|y|} \sum_i^{|y|} \log P_t(y_i|x, y_{<i})
$.
While the token-averaged reranking heuristic typically selects better translations, it is still misaligned with the final evaluation metrics used in machine translation (e.g. BLEU) \cite{papineni-etal-2002-bleu}.
In this work, we investigate whether one can train a model to predict the target evaluation metric and use it to rerank candidate translations, without adding external data or extra parameters to the reranker, in addition to the basic translation model.

\section{Method}
\label{sec:method}

We describe a simple method, BLEUR (BLEU Reranking), for training a reranker without introducing new training data or increasing the model's size.
Given a trained translation model, the main idea is to sample its translations of its own training set, compute each translation's evaluation score (e.g. BLEU), and then fine-tune a copy of the translation model to predict the evaluation score given the source and model-generated translation.
At inference time, the reranker scores each candidate translation produced by the original translation model, and selects the best option.
We also discuss a hybrid model that interpolates between the trained reranker's predicted score and the original translation model's average log-probability.

\paragraph{Objective}
We assume that the reranking model $P_r$ predicts a probability score $P_r(y|x)$ given the source $x$ and the candidate translation $y$.
For example, when predicting a translation's BLEU score, a prediction of 0.473 would be equivalent to predicting 47.3 BLEU.
While other rerankers optimize for relative scores \cite{lee-etal-2021-discriminative}, our experiments focus on predicting the absolute score.

To optimize the reranker, we use KL divergence as follows, where $p$ denotes the ground truth probability score:
\begin{align*}
\ell = &- \left[ p \log P_r(y|x) + (1-p) \log (1 - P_r(y|x)) \right] \\
&+ \left[ p \log p + (1-p) \log (1 - p) \right]
\end{align*}
The first row of the objective is a soft generalization of the cross-entropy loss over two classes, and the second row (whose derivative is zero) subtracts the optimal cross-entropy loss given the ground truth $p$ from the objective function.

\paragraph{Architecture}
To control the effects of model size, we use a copy of the original translation model to train the reranker.
In this work, we use encoder-decoder transformers \cite{Vaswani2017AttentionIA}.
We implement the reranker's probability score by taking $\vec{y}_{eos}$, the transformer's decoder-side $d$-dimensional representation of the last token in the sequence (\texttt{EOS}), passing it through a $d \times 2$ linear layer, and then applying softmax between the two logits.
The first probability is considered the predicted probability $P_r(y|x)$ and the second is its complement.

\paragraph{Training Data}
To train the reranker, we expand the original training set by producing additional model-generated translations.
Specifically, we collect a set of translations $Y$ for each source sentence $x$ using the following decoding algorithms: (1) the ground truth, (2) greedy decoding, (3) beam search (retaining only the top beam),
(4) a sample of $n$ translations using top-$k$ sampling \cite{fan-etal-2018-hierarchical}, and (5) a sample of $n$ translations using top-$p$ (nucleus) sampling \cite{Holtzman2020TheCC}.
This may amount to a maximum of $2n+3$ translations per source, but there are also some overlaps in practice.
Each translation $y \in Y$ is then scored against the reference using the evaluation metric (e.g. BLEU), producing the target score $p$.

\paragraph{Pretraining}
In our experiments, we initialize the reranking model using the original translation model's parameters.
In other words, we pretrain the reranker on the translation task.
The only parameters that require initialization are in the $d \times 2$ linear prediction head.

\paragraph{Hybrid Model}
We also experiment with a hybrid model that combines the translation model's probabilities with the reranking model's predictions.
Specifically, we interpolate the translator's average log-probability with the reranker's log-probability, denoting $\alpha$ as the interpolation weight:
\begin{align*}
\frac{1 - \alpha}{|y|} \sum_i \log P_t(y_i|x, y_{<i}) + \alpha \log P_r(y|x)
\end{align*}

\section{Experiment Setup}

\paragraph{Datasets}
We use the IWSLT2014 datasets \cite{Cettolo2015ReportOT} of English TED talks translated into German, Arabic, and Russian, as well as the WMT2016 Romanian to English dataset \cite{bojar-etal-2016-findings}.
All datasets are tokenized into subwords with BPE \cite{sennrich-etal-2016-neural}.

\paragraph{Model}
We use \texttt{fairseq}'s \cite{ott-etal-2019-fairseq} implementation of encoder-decoder transformers \cite{Vaswani2017AttentionIA} for translation and reranking models.
We use $6$ encoder and decoder layers, $4$ attention heads, a hidden dimension of $512$, and a feed-forward dimension of $1024$.
Vocabularies and embeddings are shared across source and target.

\paragraph{Hyperparameters}
We optimize with Adam \cite{Kingma2015AdamAM}, using a learning rate of $5e-4$ for translation models and $1e-4$ for rerankers, warmup of $1000$ steps, weight decay of $1e-4$, dropout of $0.3$, and for the translation models we use label smoothing of $0.1$.
We train each model until convergence: a maximum of 40k steps for translation models, and 60k steps for rerankers.
We set the batch size according to a maximum of 75k tokens for translation models, and 100k tokens for each reranker.

\paragraph{Model Selection}
To select the best checkpoint, we evaluate all models with BLEU \cite{papineni-etal-2002-bleu} on the development set.
Reranker evaluation simulates the inference environment by providing the reranker with the top $k=5$ beam search candidates, and then computing BLEU for the top-scoring solution.
Translation models are evaluated every epoch, and rerankers are checked every $200$ steps.
For the hybrid reranker model, we also tune the reranker's interpolation weight during the validation process using the values $0.1, 0.2, \ldots, 0.9$.

\paragraph{Reranking Data}
For every translation dataset we generate new reranker scoring datasets as described in Section~\ref{sec:method}.
We always use $k=10$ for top-$k$ sampling and $p=0.99$ for nucleus (top-$p$) sampling.
A temperature of $\tau=3$ is used in top-$k$ sampling to increase output diversity.
We experiment with 5 different dataset sizes by controlling $n$, the number of translations from the stochastic generation algorithms; specifically, we use $n=1, 3, 10, 32, 100$ to approximate the powers of $\sqrt{10}$.

\paragraph{Inference}
The translation candidates in all of our experiments are generated by beam search with a beam size of $b=5$.

\section{Results}

\begin{figure*}[t]
    \small
    \centering
    \begin{subfigure}[b]{.45\linewidth}
        \centering
        \includegraphics[width=\linewidth]{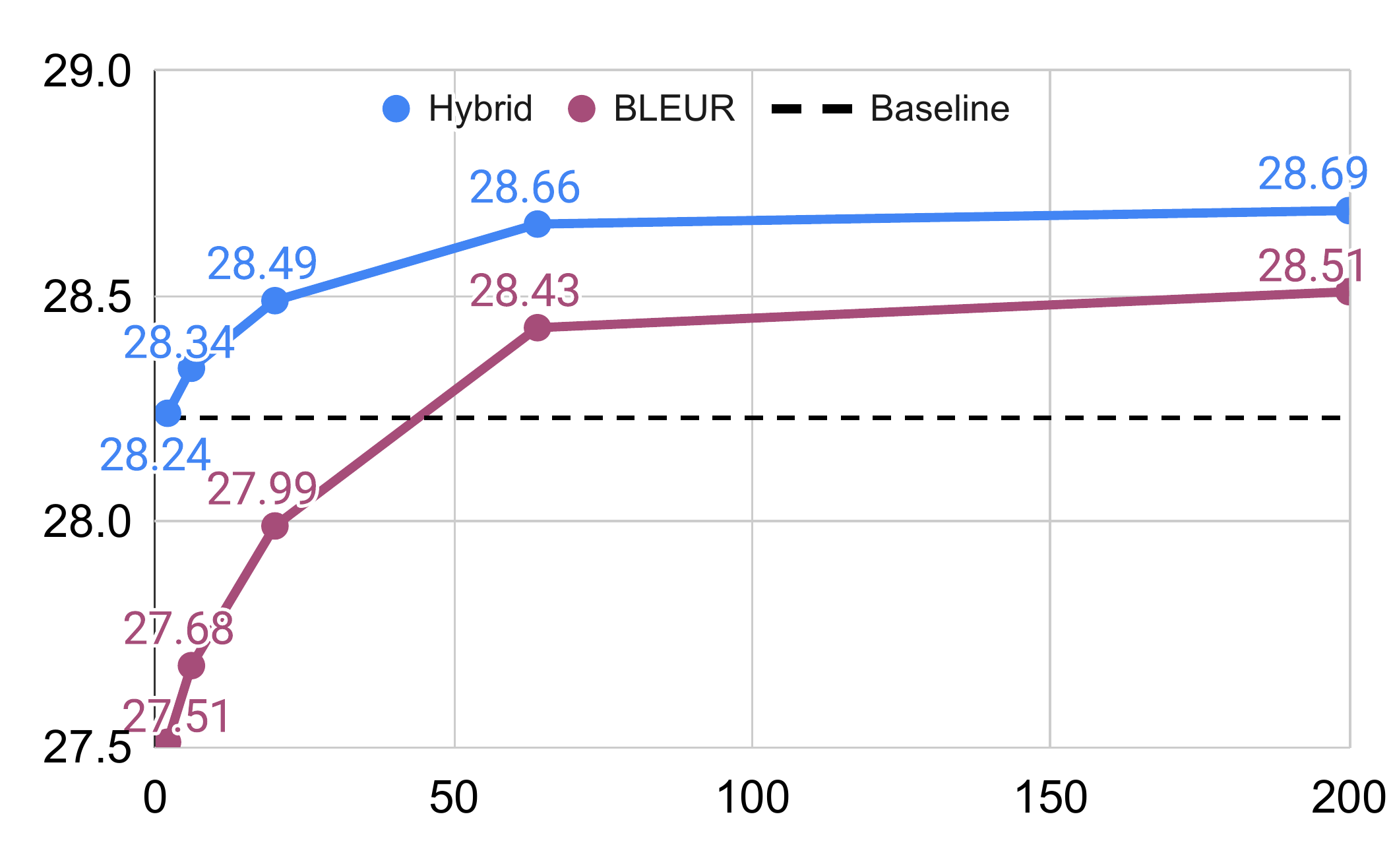} 
        \caption{IWSLT'14 EN-DE} 
        \label{fig:de}
    \end{subfigure}
    \begin{subfigure}[b]{.45\linewidth}
        \centering
        \includegraphics[width=\linewidth]{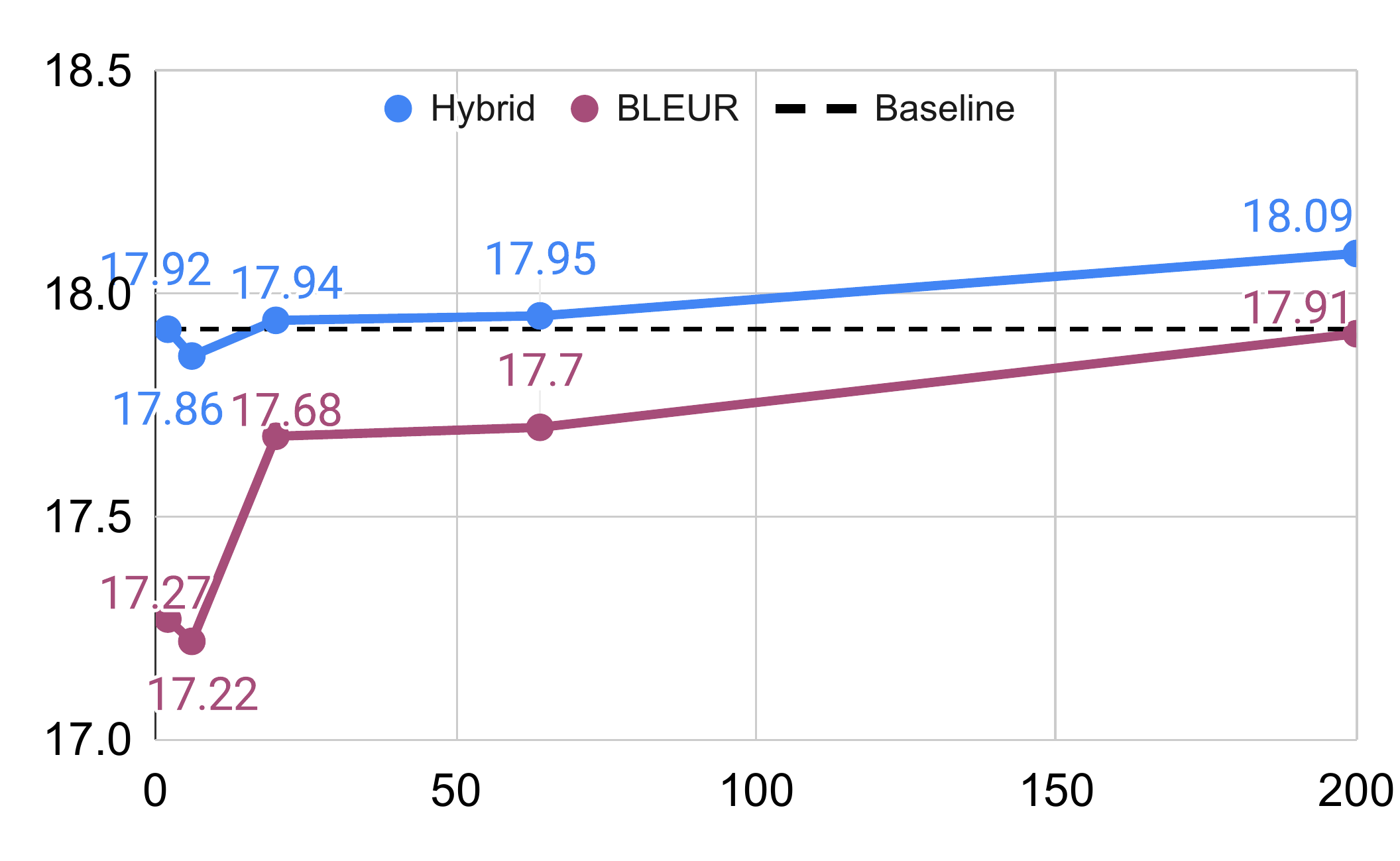} 
        \caption{IWSLT'14 EN-RU} 
        \label{fig:ru}
    \end{subfigure}
    \begin{subfigure}[b]{.45\linewidth}
        \centering
        \includegraphics[width=\linewidth]{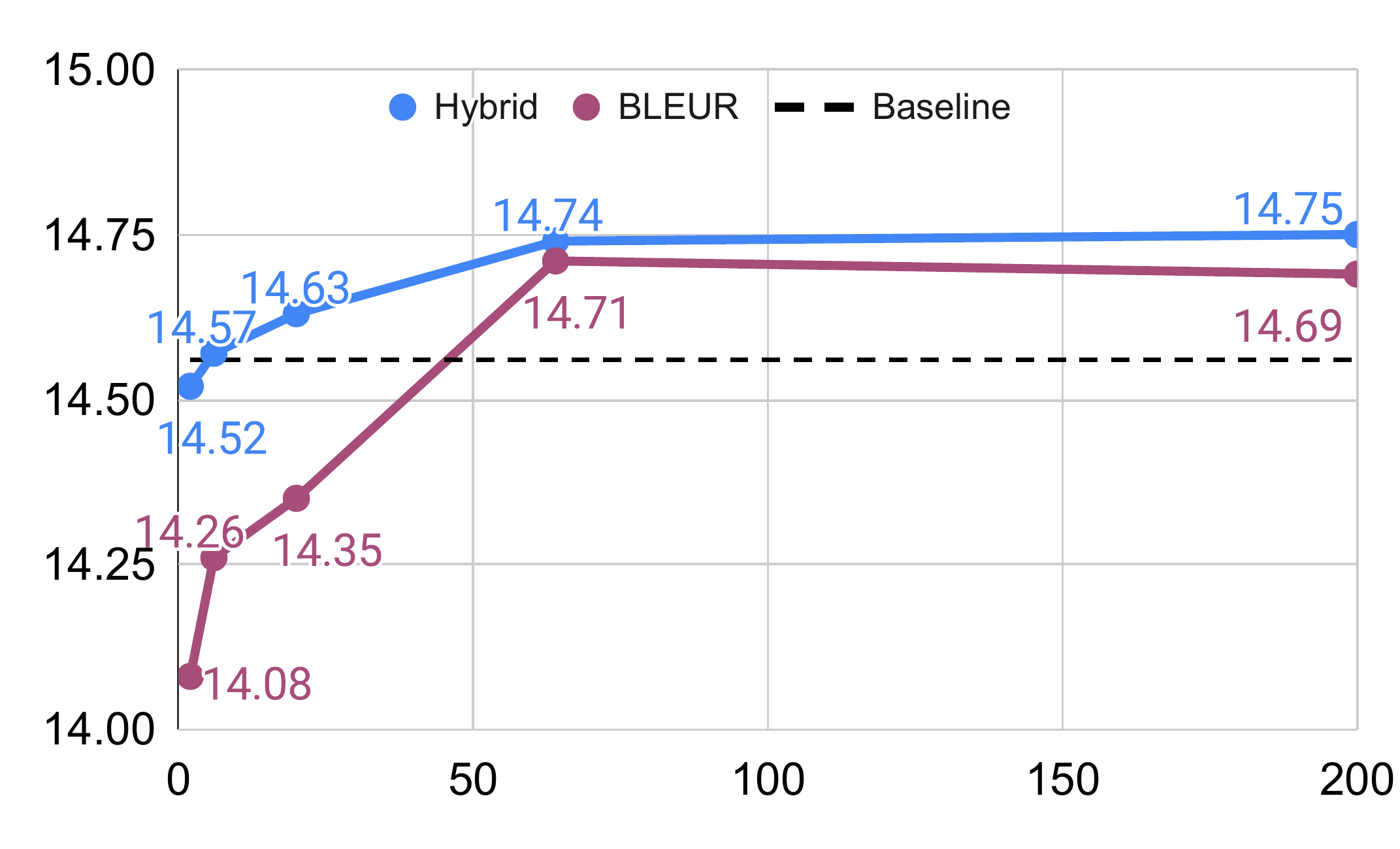} 
        \caption{IWSLT'14 EN-AR} 
        \label{fig:ar}
    \end{subfigure}
    \begin{subfigure}[b]{.45\linewidth}
        \centering
        \includegraphics[width=\linewidth]{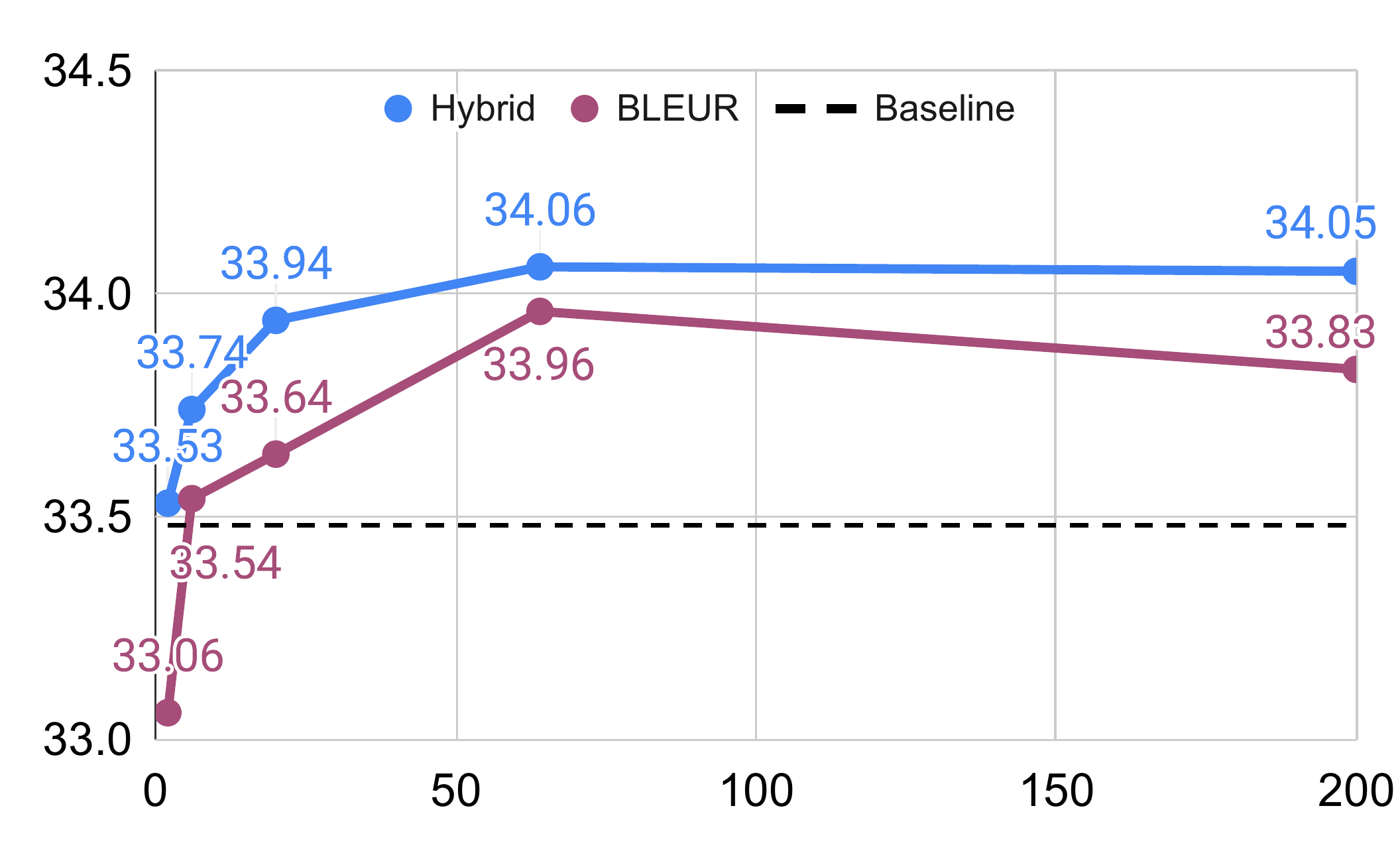} 
        \caption{WMT'16 RO-EN} 
        \label{fig:ro}
    \end{subfigure}
    \caption{Test results (BLEU) for every translation dataset, using either the baseline reranker (token-averaged log-probability), BLEUR, or the hybrid reranker.
    The horizontal axis is the number of randomly sampled translations per source sentence (using both top-k and nucleus sampling), reflecting the size of the trained reranker's dataset.}
    \label{fig:results}
    \vspace{-5pt}
\end{figure*}

Figure ~\ref{fig:results} shows the performance of the baseline model and reranker, compared to the proposed BLEUR reranking model and their interpolation, the hybrid model.
When compared to the baseline, performance is largely comparable, with some small improvements of up to 0.58 BLEU.
Individually, BLEUR's performance increases within a typical range of $1$ BLEU as the reranking training set grows, reaching saturation at around $64$ ($n=32$) random samples per source.
Furthermore, we observe that the hybrid model consistently reaches better performance compared to BLEUR, and was able to outperform the baseline model with a smaller training set.
Overall, these improvements are minor when compared to the full reranking potential as seen in Table~\ref{tab:oracle}.

\begin{figure}[t]
    \centering
    \includegraphics[width=0.9\linewidth]{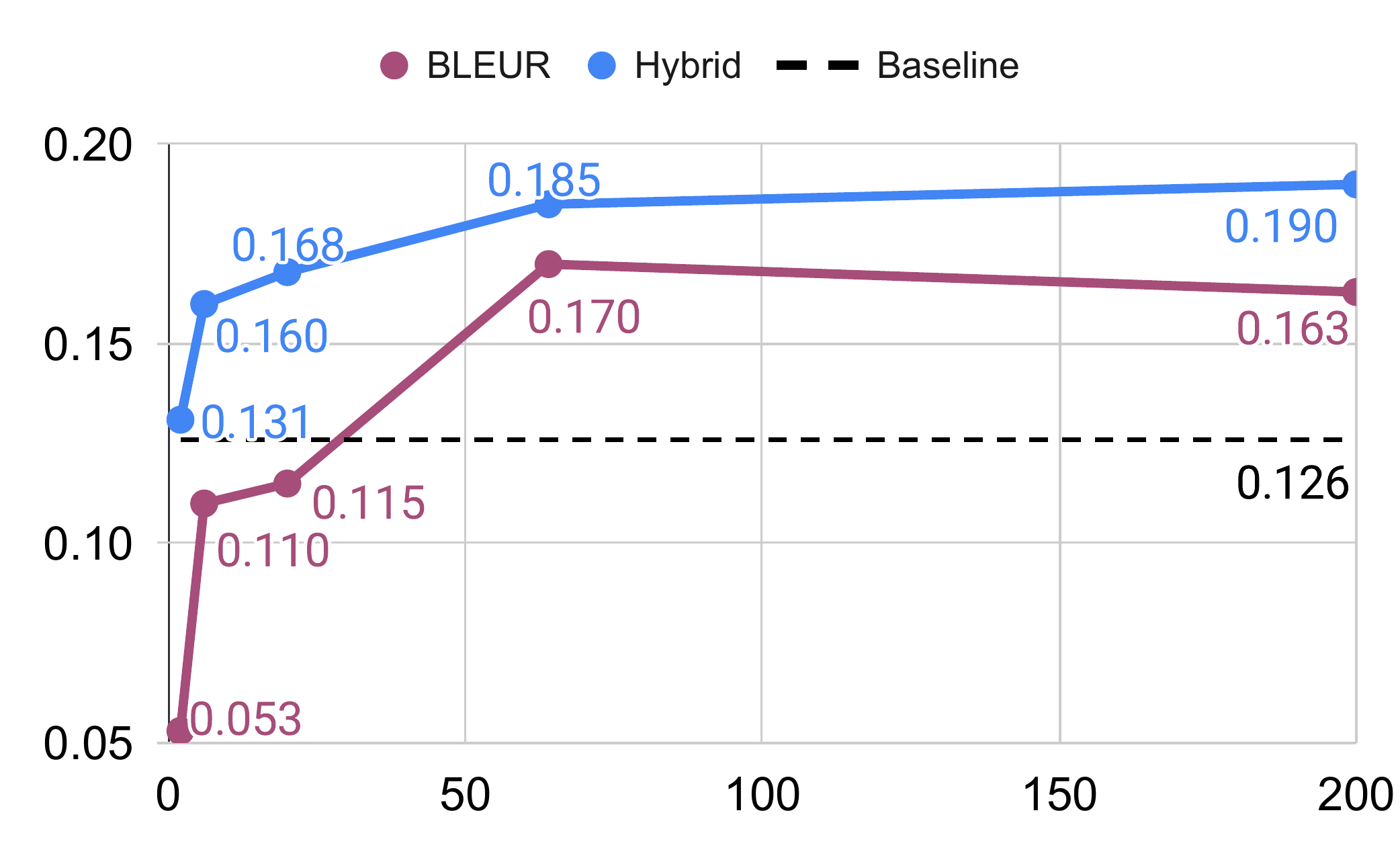} 
    \caption{Average Pearson correlation between reranker scores and real BLEU scores of beam search candidates in WMT’16 RO-EN, as a function of the number of randomly sampled translation per source sentence.}
    \label{fig:pearson}
    \vspace{-10pt}
\end{figure}

Since the main performance metric only reflects the reranker's ability to assign the top score to a good candidate, we further analyze the correlations between the reranker's scores and the actual BLEU.
Specifically, we compute Pearson correlation for each example's beam search candidates, and average it across the test set.
Figure~\ref{fig:pearson} shows that, with sufficient training examples, the reranker's scores correlate better with the target BLEU scores than the baseline, demonstrating the potential of our simple approach as a baseline in future studies. 

It would seems that the performance demonstrated in our setting is somewhat underwhelming when compared to recent literature on BLEU-based reranking \cite{lee-etal-2021-discriminative, bhattacharyya-etal-2021-energy}.
However, those experiments are not comparable to ours as they train rerankers with additional data and parameters that were not available to the original translation models.

\begin{table}[t]
\small
\centering
\begin{tabular}{@{}lrrrr@{}}
\toprule
 & \multicolumn{3}{c}{\textbf{IWSLT'14}} & \textbf{WMT'16} \\
 & \textbf{EN-DE} & \textbf{EN-AR} & \textbf{EN-RU} & \textbf{RO-EN} \\
\midrule
\textbf{Baseline} & $28.23$ & $14.56$ & $17.92$ & $33.48$ \\
\textbf{BLEUR} & $28.51$ & $14.71$ & $17.91$ & $33.96$ \\
\textbf{Hybrid} & $28.69$ & $14.75$ & $18.09$ & $34.06$\\
\textbf{Oracle} & $32.93$ & $17.71$ & $21.25$ & $37.50$ \\
\bottomrule
\end{tabular}
\caption{Reranking test results (BLEU) when given the maximal amount of training data ($n=200$), compared to the baseline reranker and the oracle reranker.}
\label{tab:oracle}
\vspace{-5pt}
\end{table}

\section{Related Work}

Recent work on reranking in machine translation includes both absolute scorers and relative rerankers.
\citet{fernandes-etal-2022-quality} propose ensembling a variety of pretrained quality estimation metrics (including the translation model's log-probability) to predict an absolute score for each candidate.
This method effectively introduces more data and parameters into the system by running inference on external pretrained language models.
In addition, our work assumes only a simple setting with 5 beam search candidates, whereas \citet{fernandes-etal-2022-quality} experiment with up to 200 candidates, as well as slower decoding methods such as MBR \cite{eikema-aziz-2020-map}; \citet{Zhang2022RMBR} also explore reranking in an MBR decoding setting.

Both energy-based reranking \cite{bhattacharyya-etal-2021-energy} and discriminative reranking \cite{lee-etal-2021-discriminative} use a comparative softmax to induce relative scores among the candidates.
Both works use pretrained models that are larger than the original translation model to train the reranker, thus introducing more parameters and monolingual data into the system.
\citet{lee-etal-2021-discriminative} also use back-translated data to train the reranker; when this data is ablated, gains from reranking are more modest.
In contrast, our experiments control for additional pretraining, data, and parameters, and examine the effect of the reranking mechanism alone.

\section{Limitations}

Our work provides a simple baselines for research on machine translation reranking.
Although we raise the hypothesis that previous studies were able to reach significant gains by injecting external factors into the rerankers, we do not attempt to prove this rigorously; that would require running additional experiments on \textit{relative} reranking models in our controlled setting.
Moreover, our work is limited to relatively small datasets, and investigating our approach using another evaluation metric (e.g. COMET), was also left for future work.

\bibliographystyle{acl_natbib}
\bibliography{anthology,custom}

\begin{thebibliography}{14}
\expandafter\ifx\csname natexlab\endcsname\relax\def\natexlab#1{#1}\fi

\bibitem[{Bhattacharyya et~al.(2021)Bhattacharyya, Rooshenas, Naskar, Sun,
  Iyyer, and McCallum}]{bhattacharyya-etal-2021-energy}
Sumanta Bhattacharyya, Amirmohammad Rooshenas, Subhajit Naskar, Simeng Sun,
  Mohit Iyyer, and Andrew McCallum. 2021.
\newblock \href {https://doi.org/10.18653/v1/2021.acl-long.349} {Energy-based
  reranking: Improving neural machine translation using energy-based models}.
\newblock In \emph{Proceedings of the 59th Annual Meeting of the Association
  for Computational Linguistics and the 11th International Joint Conference on
  Natural Language Processing (Volume 1: Long Papers)}, pages 4528--4537,
  Online. Association for Computational Linguistics.

\bibitem[{Bojar et~al.(2016)Bojar, Chatterjee, Federmann, Graham, Haddow, Huck,
  Jimeno~Yepes, Koehn, Logacheva, Monz, Negri, N{\'e}v{\'e}ol, Neves, Popel,
  Post, Rubino, Scarton, Specia, Turchi, Verspoor, and
  Zampieri}]{bojar-etal-2016-findings}
Ond{\v{r}}ej Bojar, Rajen Chatterjee, Christian Federmann, Yvette Graham, Barry
  Haddow, Matthias Huck, Antonio Jimeno~Yepes, Philipp Koehn, Varvara
  Logacheva, Christof Monz, Matteo Negri, Aur{\'e}lie N{\'e}v{\'e}ol, Mariana
  Neves, Martin Popel, Matt Post, Raphael Rubino, Carolina Scarton, Lucia
  Specia, Marco Turchi, Karin Verspoor, and Marcos Zampieri. 2016.
\newblock \href {https://doi.org/10.18653/v1/W16-2301} {Findings of the 2016
  conference on machine translation}.
\newblock In \emph{Proceedings of the First Conference on Machine Translation:
  Volume 2, Shared Task Papers}, pages 131--198, Berlin, Germany. Association
  for Computational Linguistics.

\bibitem[{Cettolo et~al.(2015)Cettolo, Niehues, St{\"u}ker, Bentivogli, and
  Federico}]{Cettolo2015ReportOT}
M.~Cettolo, J.~Niehues, S.~St{\"u}ker, L.~Bentivogli, and Marcello Federico.
  2015.
\newblock Report on the 11th iwslt evaluation campaign, iwslt 2014.

\bibitem[{Eikema and Aziz(2020)}]{eikema-aziz-2020-map}
Bryan Eikema and Wilker Aziz. 2020.
\newblock \href {https://doi.org/10.18653/v1/2020.coling-main.398} {Is {MAP}
  decoding all you need? the inadequacy of the mode in neural machine
  translation}.
\newblock In \emph{Proceedings of the 28th International Conference on
  Computational Linguistics}, pages 4506--4520, Barcelona, Spain (Online).
  International Committee on Computational Linguistics.

\bibitem[{Fan et~al.(2018)Fan, Lewis, and Dauphin}]{fan-etal-2018-hierarchical}
Angela Fan, Mike Lewis, and Yann Dauphin. 2018.
\newblock \href {https://doi.org/10.18653/v1/P18-1082} {Hierarchical neural
  story generation}.
\newblock In \emph{Proceedings of the 56th Annual Meeting of the Association
  for Computational Linguistics (Volume 1: Long Papers)}, pages 889--898,
  Melbourne, Australia. Association for Computational Linguistics.

\bibitem[{Fernandes et~al.(2022)Fernandes, Farinhas, Rei, De~Souza, Ogayo,
  Neubig, and Martins}]{fernandes-etal-2022-quality}
Patrick Fernandes, Ant{\'o}nio Farinhas, Ricardo Rei, Jos{\'e} De~Souza, Perez
  Ogayo, Graham Neubig, and Andre Martins. 2022.
\newblock \href {https://aclanthology.org/2022.naacl-main.100} {Quality-aware
  decoding for neural machine translation}.
\newblock In \emph{Proceedings of the 2022 Conference of the North American
  Chapter of the Association for Computational Linguistics: Human Language
  Technologies}, pages 1396--1412, Seattle, United States. Association for
  Computational Linguistics.

\bibitem[{Holtzman et~al.(2020)Holtzman, Buys, Forbes, and
  Choi}]{Holtzman2020TheCC}
Ari Holtzman, Jan Buys, Maxwell Forbes, and Yejin Choi. 2020.
\newblock The curious case of neural text degeneration.
\newblock \emph{ArXiv}, abs/1904.09751.

\bibitem[{Kingma and Ba(2015)}]{Kingma2015AdamAM}
Diederik~P. Kingma and Jimmy Ba. 2015.
\newblock Adam: A method for stochastic optimization.
\newblock \emph{CoRR}, abs/1412.6980.

\bibitem[{Lee et~al.(2021)Lee, Auli, and
  Ranzato}]{lee-etal-2021-discriminative}
Ann Lee, Michael Auli, and Marc{'}Aurelio Ranzato. 2021.
\newblock \href {https://doi.org/10.18653/v1/2021.acl-long.563} {Discriminative
  reranking for neural machine translation}.
\newblock In \emph{Proceedings of the 59th Annual Meeting of the Association
  for Computational Linguistics and the 11th International Joint Conference on
  Natural Language Processing (Volume 1: Long Papers)}, pages 7250--7264,
  Online. Association for Computational Linguistics.

\bibitem[{Ott et~al.(2019)Ott, Edunov, Baevski, Fan, Gross, Ng, Grangier, and
  Auli}]{ott-etal-2019-fairseq}
Myle Ott, Sergey Edunov, Alexei Baevski, Angela Fan, Sam Gross, Nathan Ng,
  David Grangier, and Michael Auli. 2019.
\newblock \href {https://doi.org/10.18653/v1/N19-4009} {fairseq: A fast,
  extensible toolkit for sequence modeling}.
\newblock In \emph{Proceedings of the 2019 Conference of the North {A}merican
  Chapter of the Association for Computational Linguistics (Demonstrations)},
  pages 48--53, Minneapolis, Minnesota. Association for Computational
  Linguistics.

\bibitem[{Papineni et~al.(2002)Papineni, Roukos, Ward, and
  Zhu}]{papineni-etal-2002-bleu}
Kishore Papineni, Salim Roukos, Todd Ward, and Wei-Jing Zhu. 2002.
\newblock \href {https://doi.org/10.3115/1073083.1073135} {{B}leu: a method for
  automatic evaluation of machine translation}.
\newblock In \emph{Proceedings of the 40th Annual Meeting of the Association
  for Computational Linguistics}, pages 311--318, Philadelphia, Pennsylvania,
  USA. Association for Computational Linguistics.

\bibitem[{Sennrich et~al.(2016)Sennrich, Haddow, and
  Birch}]{sennrich-etal-2016-neural}
Rico Sennrich, Barry Haddow, and Alexandra Birch. 2016.
\newblock \href {https://doi.org/10.18653/v1/P16-1162} {Neural machine
  translation of rare words with subword units}.
\newblock In \emph{Proceedings of the 54th Annual Meeting of the Association
  for Computational Linguistics (Volume 1: Long Papers)}, pages 1715--1725,
  Berlin, Germany. Association for Computational Linguistics.

\bibitem[{Vaswani et~al.(2017)Vaswani, Shazeer, Parmar, Uszkoreit, Jones,
  Gomez, Kaiser, and Polosukhin}]{Vaswani2017AttentionIA}
Ashish Vaswani, Noam~M. Shazeer, Niki Parmar, Jakob Uszkoreit, Llion Jones,
  Aidan~N. Gomez, Lukasz Kaiser, and Illia Polosukhin. 2017.
\newblock Attention is all you need.
\newblock \emph{ArXiv}, abs/1706.03762.

\bibitem[{Zhang et~al.(2022)Zhang, Wan, Liu, Yang, and He}]{Zhang2022RMBR}
Yidan Zhang, Yu~Wan, Dayiheng Liu, Baosong Yang, and Zhenan He. 2022.
\newblock Rmbr: A regularized minimum bayes risk reranking framework for
  machine translation.
\newblock \emph{arXiv}, abs/2203.00201.

\end{thebibliography}

\end{document}